\documentclass[10pt,journal,compsoc]{IEEEtran}

\ifCLASSOPTIONcompsoc
  \usepackage[nocompress]{cite}
\else
  \usepackage{cite}
\fi

\usepackage{mydefs}

\usepackage{graphicx}
\usepackage{subfigure}
\usepackage{tabularx}
\usepackage{algorithm}
\usepackage{algorithmic}
\usepackage{multirow}
\usepackage{forloop}
\usepackage{calc}
\usepackage{rotating}
\usepackage[table]{xcolor}
\usepackage{enumitem}
\usepackage{xspace}
\usepackage{pgfplots}
\usepackage{tikz}
\usepackage{color}

\newcommand{\yrcite}{\cite}
\begin{document}

\title{CO2 Forest: Improved Random Forest by Continuous Optimization of Oblique Splits}

\author{Mohammad~Norouzi, Maxwell~D.~Collins, David~J.~Fleet,
  Pushmeet~Kohli
  \IEEEcompsocitemizethanks{\IEEEcompsocthanksitem M. Norouzi and
    D. J. Fleet are with the Department of Computer Science, University
    of Toronto. Email: \{norouzi,
      fleet\}@cs.toronto.edu
    \IEEEcompsocthanksitem M. D. Collins is with
    the Department of Computer Science, University of
    Wisconsin-–Madison. Email: mcollins@cs.wisc.edu
    \IEEEcompsocthanksitem P. Kohli is with Microsoft Research,
    Cambridge, UK. Email: pkohli@microsoft.com
}
}

%

\IEEEtitleabstractindextext{%
\begin{abstract}
We propose a novel algorithm for optimizing multivariate linear threshold functions as split functions of decision trees to create improved Random Forest classifiers. Standard tree induction methods resort to sampling and exhaustive search to find good univariate split functions. In contrast, our method computes a linear combination of the features at each node, and optimizes the parameters of the linear combination (oblique) split functions by adopting a variant of latent variable SVM formulation. We develop a convex-concave upper bound on the classification loss for a one-level decision tree, and optimize the bound by stochastic gradient descent at each internal node of the tree. Forests of up to 1000 Continuously Optimized Oblique (CO2) decision trees are created, which significantly outperform Random Forest with univariate splits and previous techniques for constructing oblique trees. Experimental results are reported on multi-class classification benchmarks and on Labeled Faces in the Wild (LFW) dataset.

\comment{This bound is optimized using stochastic gradient descent to provide a locally optimal configuration of the parameters.}
\comment{{
This tree induction algorithm is suboptimal since the split functions at each node of the tree are learned without considering the split functions at the subsequent (child) nodes.
}}

\end{abstract}

}

\maketitle

\IEEEdisplaynontitleabstractindextext

\IEEEpeerreviewmaketitle

\section{Introduction}

Decision trees~\cite{breiman1984classification, quinlan1986induction}
and random forests~\cite{ho1998random,Breiman01} have a long,
successful history in machine learning, in part due to their
computational efficiency and their applicability to large-scale
classification and regression tasks (\eg~see~\cite{trevor2009elements,
  AntonioJamieBook}).  A case in point is the Microsoft Kinect, where
multiple decision trees are learned on millions of training exemplars
to enable real time human pose estimation from depth
images~\cite{ShottonPami13}.  The standard algorithm for decision tree
induction grows a tree one node at a time, greedily and
recursively. The building block of this procedure is an optimization
at each internal node of the tree, which divides the training data at
that node into two subsets according to a splitting criterion, such as
Gini impurity index in {\em CART}~\cite{breiman1984classification}, or information gain in {\em
  C4.5}~\cite{quinlan1993c4}. This corresponds to optimizing a binary
{\em decision stump}, or a one-level decision tree, at each internal
node.  Most tree-based methods exploit univariate (axis-aligned) split
functions, which compare one feature dimension to a
threshold. Optimizing univariate decision stumps is straightforward
because one can exhaustively enumerate all plausible thresholds for
each feature, and thereby select the best parameters according to the
split criterion.  Conversely, univariate split functions have limited
discriminative power.


We investigate the use of a more general and powerful family of split
functions, namely, linear-combination (\aka~{\em oblique}) splits.
Such split functions comprise a multivariate linear projection of the
features followed by binary quantization. Clearly, exhaustive search
with linear hyperplanes is not feasible, and based on our preliminary
experiments, random sampling yields poor results.  Further, typical
splitting criteria for a one-level decision tree (decision stump) are
discontinuous, since small changes in split parameters may change the
assignment of data to branches of the tree.  As a consequence, split
parameters are not readily amenable to numerical optimization, so
oblique split functions have not been used widely with tree-based
methods.

This paper advocates a new building block for learning decision trees,
\ie~an algorithm for continuous optimization of oblique decision
stumps.  To this end, we introduce a continuous upper bound on the
empirical loss associated with a decision stump.  This upper bound
resembles a ramp loss, and accommodates any convex loss that is useful
for multi-class classification, regression, or structured
prediction~\cite{nowozin2011}. As explained below, the bound is the
difference of two convex terms, the optimization of which is
effectively accomplished using the Convex-Concave Procedure
of~\cite{YuilleR03}. The proposed bound resembles the bound used for
learning binary hash functions~\cite{NorouziFICML11}.

Some previous work has also considered improving the classification
accuracy of decision trees by using oblique split functions. For
example, Murthy {\em et al.}~\cite{murthy1994} proposed a method
called {\em OC1}, which yields some performance gains over {\em CART}
and {\em C4.5}. Nevertheless, individual decision trees are rarely
sufficiently powerful for many classification and regression tasks.
Indeed, the power of tree-based methods often arises from diversity
among the trees within a forest.  Not surprisingly, a key question
with optimized decision trees concerns the loss of diversity that
occurs with optimization, and hence a reduction in the effectiveness
of forests of such trees.  The {\em random forest} of Breiman seems to
achieve a good balance between optimization and randomness.

Our experimental results suggest that one can effectively optimize
oblique split functions, and the loss of diversity associated with
such optimized decision trees can be mitigated.  In particular, it is
found that when the decision stump optimization is initialized with
random forest's split functions, one can indeed construct a forest of
non-correlated decision trees.  We effectively take advantage of the
underlying non-convex optimization problem, for which a diverse set of
initial states for the optimizer yields a set of different split
functions.  Like random forests, the resulting algorithm achieves very
good performance gains as the number of trees in the ensemble
increases.

We assess the effectiveness of our tree construction algorithm by
generating up to $1000$ decision trees on nine classification benchmarks.  
Our algorithm, called CO2 forest, outperforms random
forest on all of the datasets.  It is also shown to outperform a 
baseline of OC1 trees. As a large-scale experiment, we consider the 
task of segmenting faces from the Labeled Faces in the Wild (LFW) 
dataset \cite{huang2007-lfw}. Again, our results confirm that CO2 
forest outperforms other baselines.

\comment{Hence combining optimized decision trees is typically not as
  helpful as using random decision trees.}
\comment{Further, our continuous optimization framework can be used
  within other tree-based methods too such as gradient boosted
  trees~\cite{friedman2001greedy}}
\comment{Computational efficiency and high model fidelity are
  characteristics that make decision trees and forests particularly
  suitable for large-scale machine learning tasks.}
\comment{{
These models are becoming increasingly popular in fields such as
Computer Vision due to the on-going exponential growth in information.
Computational efficiency and high model fidelity are characteristics
that make decision trees and forests particularly suitable for
large-scale machine learning tasks including regression and
classification~\cite{ShottonPami13}.  A case in point is the Microsoft
Kinect where decision trees are learned on millions of training data
points to enable real-time human pose estimation from depth
images~\cite{ShottonPami13}.

\comment{{
The last few years have seen decision
trees/forests~\cite{hunt1966experiments,breiman1984classification}
becoming increasingly popular in fields such as Computer Vision to
solve a wide range of classification and regression
problems~\cite{AntonioJamieBook}.  Computational efficiency and high
model fidelity are characteristics that make decision trees
particularly suitable for machine learning tasks that involve
large-scale datasets (Big Data). A case in point is the Microsoft
Kinect where decision trees have enabled real time human pose
estimation from depth images~\cite{ShottonPami13}.  }}


\comment{Even though decision trees have been used extensively, the method for
training them has not changed fundamentally over the years.}

Recent work has considered other splitting objectives for learning
decision trees based on specific loss
functions~\cite{NowozinICML12,JancsaryECCV12}, and some new
applications are proposed~\cite{gall2011, konukoglu2012}, but, the
recursive greedy optimization has remained at the heart of the
training procedures for tree-based models.

An intrinsic limitation of any greedy tree induction algorithm is that
when the split functions at the top levels of the tree are being
optimized, the algorithm is unaware of the split functions to be
introduced at the bottom levels later. This paper proposes a general
framework for {\em non-greedy} learning of linear combination splits
for tree-based methods. We show that jointly optimizing the split
functions at different levels of the tree, promotes cooperation
between the split nodes, resulting in more compact trees, and better
generalization power. This paper concerns classification decision
trees, but the proposed framework can be adopted for other tasks such
as regression and density estimation.

\mohammad{linear combination splits - Oblique}\\
\mohammad{little bump here}


One of our key contributions is to recognize that the decision tree
optimization problem is an instance of structured prediction with
latent variables~\cite{YuJ09}. We present a novel representation of
the decision trees, which associates a binary {\em latent decision
variable} with each split node in the tree, and we formulate an
empirical loss function on the decision random variables. Inspired by
recent advances in structured prediction\comment{with and without
latent variables}~\cite{TaskarGK03, TsochantaridisHJA04, YuJ09}, we
formulate a convex-concave upper bound on the tree's empirical
loss. This bound acts as a surrogate objective, which is optimized
using the convex concave procedure (CCCP)~\cite{YuilleR03} or
stochastic gradient descent (SGD) to find a locally optimal
configuration of the split functions.

The number of latent decision variables in our model is exponential in
the tree depth, and each gradient update step for the CCCP and SGD
requires a computational cost of $O(2^d)$, where $d$ is the tree
depth. One of our technical contributions is to show how this
complexity for SGD can be reduced to $O(d^2)$ by slight modifications
to the surrogate objective. This enables efficient learning of deep
trees.
}}

\comment{{
Not only does this representation play a key part in enabling our
method for jointly learning the split functions of the trees, but it
also points to interesting connections between decision trees and
hashing method for classification. In fact, it can be shown that with
our representation, methods like minimal loss hashing can be seen as a
collection of two level decision trees~\cite{xx}cite min loss hashing
and other related papers).  }}

\comment{{
The method proposed in the paper can be modified to jointly learn
multiple trees.
}}

\section{Related Work}

Breiman~\etal~\cite{breiman1984classification} proposed a version of
CART that employs linear combination splits, known as
CART-linear-combination (CART-LC). Murthy~\etal~\cite{heath1993,
  murthy1994} proposed OC1, a refinement of CART-LC that uses random
restarts and random perturbations to escape local minima.  The main
idea behind both algorithms is to use coordinate descent to optimize
the parameters of the oblique splits one dimension at a time. Keeping
all of the weights corresponding to an oblique decision stump fixed
except one, for each datum they compute the critical value of the
missing weight at which the datum switches its assignment to the
branches. Then, one can sort these critical values to find the optimal
value of each weight (with other weights fixed).  By performing
multiple passes over the dimensions and the data, oblique splits with
small empirical loss can be found.

By contrast, our algorithm updates all the weights simultaneously
using gradient descent. While the aforementioned algorithms 
focus mainly on optimizing the splitting criterion to minimize tree 
size, there is little promise of improved generalization. Here, by 
adopting a formulation based on the latent variable SVM~\cite{YuJ09}, 
our algorithm provides a natural means of regularizing the oblique 
split stumps, thereby improving the generalization power of the trees.


The hierarchical mixture of experts (HME)~\cite{JordanJacobs94} uses
{\em soft} splits rather than hard binary decisions to capture situations
where the transition from low to high response is gradual.  The
empirical loss associated with HME is a smooth function of the unknown
parameters and hence numerical optimization is feasible. The main
drawback of HME concerns inference.  That is, multiple paths along 
the tree should be explored during inference, which reduces the 
efficiency of the classifier.

Our work builds upon random forest~\cite{Breiman01}. Random forest
combines bootstrap aggregating (bagging)~\cite{bagging} and the random
selection of features~\cite{ho1998random} to construct an ensemble of
non-correlated decision trees. The method is used widely for
classification and regression tasks, and research still investigates
its theoretical characteristics~\cite{denil2013narrowing}. Building on
random forest, we also grow each tree using a bootstrapped version of
the training dataset. The main difference is the way the split
functions are selected. Training random forest is generally faster
than using our optimized oblique trees, and because random forest uses
univariate splits, classification with the same number of trees is
often faster.  Nevertheless, we often achieve similar accuracy with
many fewer trees, and depending on the application, the gain in
classification performance is clearly worth the computational
overhead.

There also exist boosting based techniques for creating ensembles of 
decision trees~\cite{friedman2001greedy,zhu2009}. A key benefit of 
random forest over boosting is that it allows for faster training as 
the decision trees can be trained in parallel. In our experiments we
usually train $30$ trees in parallel on a multicore machine. Nevertheless, 
it is interesting to combine boosting techniques with our oblique trees, 
and we leave this to future work.


Menze~\etal~\cite{menze2011} also consider a variant of oblique random
forest. At each internal node they find an optimal split function
using either ridge regression or linear discriminant analysis.  Like
other previous work~\cite{tibshirani2007, bennett1998}, the technique 
of~\cite{menze2011} is only conveniently applicable to binary 
classification tasks. A big challenge in a multi-class setting is solving 
the combinatorial assignment of labels to the two leaves. In contrast 
to \cite{menze2011}, our technique is more general, and allows for 
optimization of multi-class classification and regression loss functions.

Rota Bul\'{o} \& Kontschieder~\cite{rota2014} recently proposed the use 
of multi-layer neural nets as split functions at internal nodes.  While 
extremely powerful, the resulting decision trees lose their computational 
simplicity during training and testing.  Further, it may be difficult to 
produce the required diversity among trees in a forest.  This paper 
explores the middle ground, with a simple, yet effective class of linear 
multi-variate split functions.  That said, note that the formulation
of the upper bound used to optimize empirical loss in this paper can
be extended to optimize other non-linear split functions, including 
neural nets (\eg \cite{NorouziFSNIPS12}).


\section{Preliminaries}

\comment{Moreover, we focus on decision trees with linear combination
  splits (\aka~{\em oblique} and multivariate splits).  We only
  consider real valued input features; categorical attributes are
  beyond the scope of our framework.}

\comment{For instance, $\x$
can represent an image patch, and $\y$ can be an object label.}

\comment{{
To keep the exposition simple, we concentrate on classification
decision trees.  A decision tree classifier with $\m$ internal (split)
nodes, and $\m+1$ leaf (terminal) nodes\footnote{In a binary tree the
  number of leaves is always one more than the number of internal
  (non-leave) nodes.}  classifies an input $\x$ in the following
manner: starting from the root, each internal node $i \in \{1, \ldots,
m\}$ performs a binary test, which involves evaluating a node-specific
split function $\test_i(\x): \Real^p \to \{-1, +1\}$ on $\x$. If
$\test_i(\x)$ is $-1$, then $\x$ is forwarded to the left child of
node $i$. Otherwise, it is forwarded to the right child. Following a
path down the tree, $\x$ reaches a leaf node $j \in \{0, \ldots,
m\}$. We store a constant parameter vector at each leaf denoted
$\leaf_j \in \Real^k$, which represents unnormalized predictive
log-probabilities in the classification case. Assuming that $\x$
terminates at leaf $j$, the predictive probability of a class label
$\alpha \in \{1,\ldots,k\}$, denoted $p(\y = \alpha\mid j)$, is
estimated by normalizing the exponential of the $\th{\alpha}$ element
of $\leaf_j$, via a softmax function:
}}

For ease of exposition, this paper is focused on binary classification 
trees, with $\m$ internal (split) nodes, and $\m+1$ leaf (terminal) 
nodes.\footnote{In a binary tree the number of leaves is always one 
more than the number of internal (non-leaf) nodes.}  
An input, $\x \in \Real^\p$, is directed from the root of the tree down 
through internal nodes to a leaf node, which specifies a distribution 
over $k$ class labels.

Each internal node, indexed by $i \in \{1, \ldots, m\}$, performs a 
binary test by evaluating a node-specific split function $\test_i(\x):
\Real^p \to \{-1, +1\}$.  If $\test_i(\x)$ evaluates to $-1$, then $\x$ 
is directed to the left child of node $i$. Otherwise, $\x$ is directed 
to the right child.  And so on down the tree.
Each split function $\test_i(\cdot)$, parametrized by a weight vector 
$\w_i$, is assumed to be a linear threshold function of the form 
$\test_i(\x) = \sign(\trans{\w_i} \x)$. We incorporate an offset parameter 
to obtain split functions of the form $\sign(\trans{\w_i} \x - b_i)$ by 
using homogeneous coordinates (\ie~by appending a constant ``$-1$'' to 
the end of the input feature vector).

Each leaf node, indexed by $j \in \{0, \ldots, m\}$, specifies 
a conditional probability distribution over class labels, 
$l \in \{1,\ldots,k\}$, denoted $p(\y = l \mid j)$.  These distributions 
are parameterized in terms of a vector of unnormalized predictive 
log-probabilities, denoted $\leaf_j \in \Real^k$, and a conventional 
softmax function; \ie
\begin{equation}
p(\y = l\mid j)
~=~ \frac{\exp \left\{ \leaf_{j[l]} \right\}}{\sum_{\alpha=1}^k\exp \left\{ \leaf_{j[\alpha]} \right\}}~,
\comment{= \frac{\exp \left\{ \leaf_{j[l]} \right\}}{\sum_{\alpha=1}^k\exp \left\{ \leaf_{j[\alpha]} \right\}}}
\label{eq:tree-pred-softmax}
\end{equation}
where $\vec{v}_{[\alpha]}$ denotes the $\alpha^\mathrm{th}$ element of
vector $\vec{v}$.

The parameters of the tree comprise the $m$ internal weight vectors, 
each of dimension $p+1$, and the $m+1$ vectors of unnormalized 
log-probabilities, one for each leaf node, \ie~$\{\w_i\}_{i=1}^m$ and 
$\{\leaf_j\}_{j=0}^m$.
Given a dataset of input-output pairs, $\Dset \equiv \{ \x_z, \y_z
\}_{z=1}^n$, where $\y_z \in \{1, \ldots, k \}$ is the ground truth
class label associated with input $\x_z \in \Real^\p$, we wish to find
a joint configuration of oblique splits $\{ \w_i \}_{i=1}^m$ and leaf
parameters $\{ \leaf_j \}_{j=0}^{m}$ that minimize some measure of
misclassification loss on the training set.  Joint optimization of the
split functions and leaf parameters according to a global objective
is, however, known to be extremely challenging~\cite{HyafilR76} due to
the discrete and sequential nature of the decisions within the tree.  

To cope with the discontinuous objective caused by discrete split
functions, we propose a smooth upper bound on the empirical loss, with
which one can effectively learn a diverse collection of trees with
oblique split functions.  We apply this approach to the optimization
of split functions of internal nodes within the context of a top-down
greedy induction procedure, one in which each internal node is treated
as an independent one-level decision stump. \comment{(\ie~a single
  internal node with two leaves)} The split functions of the tree are
optimized one node at a time, in a greedy fashion as one traverses the
tree, breadth first, from the root downward.  The procedure terminates
when a desired tree depth is reached, or when some other stopping
criterion is met.  While we focus here on the optimization of a single
stump, the formulation can be generalized to optimize entire trees.

\comment{{
Given a dataset of input output pairs $\Dset \equiv \{ \x_z, \y_z
\}_{z=1}^n$, where $\x_z \in \Real^\p$ denotes a data point, and $\y_z
\in \{1, \ldots, k \}$ is a corresponding class label, we would like
to find a joint configuration of linear combination splits $\{ \w_i
\}_{i=1}^m$ and leaf parameters $\{ \leaf_j \}_{j=0}^{m}$ minimizing
some measure of misclassification loss on the training set. However,
the problem of jointly optimizing the split functions and leaf
parameters according to a global objective is
challenging~\cite{HyafilR76} due to discrete and sequential nature of
the decisions in a decision tree. Instead, we follow the common
top-down greedy tree induction procedure, in which at every internal
node, starting from the root, we assume that this node is going to be
a node followed by two leaves; no more internal nodes will
follow. Then, we optimize a level-one decision tree, and once the
decision stump is optimized, the two leaves are replaced with two
internal nodes. we further optimize two more decision stumps given
their corresponding subsets of the data, and the procedure is
terminated when a desired tree depth is reached or some other stopping
criteria are met. Below, we describe our strategy for optimizing
oblique decision stumps, which acts as the key building block of greedy
oblique decision trees and forests.
}}



\comment{When $\w_i$'s are sparse, \ie~each $\vec{w}_i$ has only a few nonzero elements, the
split functions can be evaluated very efficiency since only a sparse
dot product between the parameter vector and the features needs to be
computed.}

\section{Continuous Optimization of Oblique (CO2) Decision Stumps}

A binary decision stump is parameterized by a weight vector $\w$, and 
two vectors of unnormalized log-probabilities for the two leaf nodes, 
$\leaf_0$ and $\leaf_1$.  The stump's loss function comprises two terms, 
one for each leaf, denoted $\ell(\leaf_0, \y)$ and $\ell(\leaf_1, \y)$,
where $\ell : \Real^k \times \{1, \ldots, k\} \to \Real^+$.  They measure 
the discrepancy between the label $y$ and the distributions 
parameterized by $\leaf_0$ and $\leaf_1$.  The binary test 
at the root of the stump acts as a gating function to select a leaf, 
and hence its associated loss.  The empirical loss for the stump, \ie~the 
sum of the loss over the training set $\cal{D}$, is defined as
\begin{equation}
\begin{aligned}
\calL&\left(\w, \leaf_{0,1} ; \Dset \right) ~= \\
&\sumxy \mathbbm{1}(\trans{\w}\x < 0) \,\ell(\leaf_0, \y)+\mathbbm{1}(\trans{\w}\x \ge 0) \,\ell(\leaf_1, \y)~,
\end{aligned}
\label{eq:emp-risk-dtree}
\end{equation}
where $\mathbbm{1}(\cdot)$ is the usual indicator function. 
Given the softmax model of \eqref{eq:tree-pred-softmax},
the log loss takes the form
\begin{equation}
\ell_{\log}(\leaf, \y) \, =\,  
-\leaf_{[\y]} 
+ \log \left\{ \sum\nolimits_{\alpha=1}^k\exp \left\{ \leaf_{[\alpha]} \right\} \right\}~.
\end{equation}

Regarding this formulation, we note that the parameters which minimize
\eqref{eq:emp-risk-dtree} with the log loss, $\ell_{\log}$, are those
that maximize information gain.  One can prove this with
straightforward algebraic manipulation of \eqref{eq:emp-risk-dtree},
recognizing that the $\leaf_0$ and $\leaf_1$ that minimize
\eqref{eq:emp-risk-dtree}, given any $\w$, are the empirical class
log-probabilities at the leaves.

We also note that the framework outlined below accommodates other 
loss functions that are convex in $\leaf$. For instance, for 
regression tasks where $\vec{y} \in \Real^k$, one can use squared loss,
\begin{equation}
\ell_{\text{sqr}}(\leaf, \vec{y}) \, =\,  \lVert \leaf - \vec{y}\rVert_2^2~~.
\end{equation}

As mentioned already above, it is also important to note that
empirical loss, $\calL(\w,\leaf_{0,1} ; \Dset)$, is a discontinuous
function of $\w$.  As a consequence, optimization of $\calL$ with 
respect to $\w$ is very challenging.  Our approach, outlined in detail 
below, is to instead optimize a continuous upper bound on empirical loss.  
This bound is closely related to formulations of binary SVM and 
logistic regression classification.  In the case of binary classification, 
the assignment of class labels to each side of the hyperplane, \ie~the 
parameters $\leaf_{0}$ and $\leaf_{1}$, are pre-specified.  In contrast, 
a decision stump with a large numbers of labels entails joint optimization 
of both the assignment of the labels to the leaves and the hyperplane 
parameters.

\comment{
\subsection{Connection with Latent SVM Models}
Before presenting upper bound on empirical loss, we point out the
connection between the problem of learning decision stumps and latent
variable SVMs.

 introducing the
surrogate A tree classifier alternates between two atomic operations:
binary test, and tree navigation.  At each split node, a binary test
is performed, and based on its result, one either navigates to the
left or right subtree, where more binary tests are performed.  Our
first observation is that decoupling the binary tests and tree
navigation steps simplifies the formulation, and connects decision
trees and latent variable models. For now, let's ignore the
hierarchical and sequential dependencies between the binary tests in a
decision tree. The binary test results at every split node can be
precomputed by $\sign(\W\x) \in \Hset \equiv \{-1, 1\}^\m$, where
$\sign$ is element wise sign function.  Given an arbitrary $\m$-bit
latent decision vector $\h \in \Hset$, a {\em tree navigation
  function} $f(\h)$ determines based on the bits in $\h$ which leaf
node will be reached.  The function $f: \Hset \to \Hoompo$, where
$\Hoompo$ denotes a $1$-of-$(\m\!+\!1)$ encoding, returns an indicator
vector corresponding to a leaf. \figref{fig:fexample} depicts a few
examples of $f(\cdot)$ for a tree of depth $\depth=2$ with $\m=3$
split nodes, and $4$ leaf nodes.

\comment{
\begin{figure}
\begin{center}
\tikzset{
  treenode/.style = {align=center, inner sep=0pt, text centered,
    font=\sffamily},
  arn_n/.style = {treenode, circle, white, font=\sffamily\bfseries, draw=black,
    fill=black, text width=1.4em,solid},
  arn_r/.style = {treenode, circle, black, draw=black, 
    text width=1.4em, very thick,solid},
  arn_r2/.style = {treenode, circle, red, draw=gray, 
    text width=1.4em, very thick,solid},
  arn_x/.style = {treenode, rectangle, draw=black,
    minimum width=0.5em, minimum height=0.5em,solid},
    emph/.style={ solid, very thick },
    norm/.style={ dashed, thin}
}

\begin{tikzpicture}[->,>=stealth',level/.style={sibling distance = 2.5cm/#1,
  level distance = 1.5cm}] 
\begin{scope}[scale = .6]
\node [arn_n,label={$h_1$}] {+1}
    child{ node [arn_n,label={$h_2$}] {-1}  
           child{ node [arn_r2, label=below:$\leaf_1$] {} edge from parent[emph]}
	child{ node [arn_r2, label=below:$\leaf_2$] {} edge from parent[norm]}
           edge from parent[norm]	
    }
    child{ node [arn_n,label={$h_3$}] {+1}
	child{ node [arn_r2, label=below:$\leaf_3$] {} edge from parent[norm]}
	child{ node [arn_r, label=below:$\leaf_4$] {} edge from parent[emph]}
           edge from parent[emph]
    }
; 
\end{scope}

\end{tikzpicture}~\hspace{.5cm}~
\begin{tikzpicture}[->,>=stealth',level/.style={sibling distance = 2.5cm/#1,
  level distance = 1.5cm}] 
\begin{scope}[scale = .7]
\node [arn_n,label={$h_1$}] {-1}
    child{ node [arn_n,label={$h_2$}] {+1}  
           child{ node [arn_r2, label=below:$\leaf_1$] {} edge from parent[norm]}
	child{ node [arn_r, label=below:$\leaf_2$] {} edge from parent[emph]}
           edge from parent[emph]	
    }
    child{ node [arn_n,label={$h_3$}] {+1}
	child{ node [arn_r2, label=below:$\leaf_3$] {} edge from parent[norm]}
	child{ node [arn_r2, label=below:$\leaf_4$] {} edge from parent[emph]}
           edge from parent[norm]
    }
; 
\end{scope}
\end{tikzpicture}
\small
\begin{tabular}{c}(left) $
  f\left(\left[\begin{array}{@{}c@{}}+1\\-1\\+1\end{array}\right]\right)
  =
  f\left(\left[\begin{array}{@{}c@{}}\,+1\,\\+1\\+1\end{array}\right]\right)
  = \left[\begin{array}{@{}c@{}}\,0\,\\0\\0\\1\end{array}\right]$
  ,\\(right) $f\left(\left[\begin{array}{@{}c@{}}
      -1\\+1\\+1 \end{array}\right]\right) =
  f\left(\left[\begin{array}{@{}c@{}}
      -1\\+1\\-1 \end{array}\right]\right) =
  \left[\begin{array}{@{}c@{}} \,0\,\\1\\0\\0 \end{array}\right].  $
\end{tabular}
\vspace*{-.3cm}
\end{center}
\caption{\footnotesize Some examples of latent decision variables and
  their corresponding values of the navigation function.  }

\label{fig:fexample}
\vspace*{-.3cm}
\end{figure}
}

\mohammad{get rid of $1/n$ and subscript $z$}
By definitions of $f(\cdot)$ and $\Leaves$, the matrix vector product
$\trans{\Leaves} f(\h)$ represents the leaf parameter chosen according
to the binary decisions encoded in $\h$. Accordingly, the predicted
parameters by $T$ for an input $\x$ can be re-expressed as
$T(\x ; \W, \Leaves) = \trans{\Leaves} f(\sign(\W \x))$,
and empirical loss can be expressed succinctly as
\begin{equation}
\calL (\W, \Leaves ; \Dset)
~=~ \sumxy \loss \bigl(\trans{\Leaves} f\big(\sign(\W \x)\big), \y \bigr)~.
\label{eq:dtree-emploss}
\end{equation}
Clearly, for computing $\trans{\Leaves}f(\sign(\W \x))$ one does not
need to evaluate all of the $\m$ binary tests first, and then navigate
the tree by evaluating $f(\cdot)$. Instead, the standard tree
prediction algorithm can be used to perform only the tests
associated with the path from the root to a final leaf. However, as
becomes clear below, the notion of $\m$-bit latent decisions play an
important role in our learning algorithm, although, we try our best to
avoid $O(\m)$ computation required for processing an $m$-bit binary
vector as $m$ is $O(2^d)$.

As another key element, our algorithm hinges on the connection between
hyperplane based binary tests, and structured prediction. This
connection has been previously pointed out by Norouzi \& Fleet
\yrcite{NorouziFICML11} for learning hyperplane based hash
functions. Motivated by their approach, we re-express the binary tests
in a decision tree as a form of structured prediction:
\begin{equation}
\sign(\W \x) = \argmax{\h \in \Hset} \left\{\trans{\h} W \x \right\}\,,
\label{eq:sign-as-max}
\end{equation}
where $\Hset \equiv \{-1, 1\}^\m$.  Binary tests implicitly map an
input $\vec{x}$ to a discrete binary vector $\h$ by maximizing the
inner product between $\h$ and $\W\x$. Here, $\trans{\h} W \x$ is the
score function, which is often expressed using a joint feature space
on $\h$ and $\x$ as $\trans{\vec{w}}\phi(\h, \x)$. In our application,
$\phi(\h, \x) = \operatorname{vec}{(\h \trans{\x})}$, and $\w =
\operatorname{vec}{(\W)}$.

To see the connection between decision tree learning and structured
prediction, we re-express empirical loss in a form similar to loss
functions for latent structured prediction:
\begin{equation}
\begin{aligned}
\calL (\W, \Leaves ; \Dset) &=~
\hspace{-.1cm}\sum_{(\x,\y) \in \Dset}\hspace{-.1cm}
\loss\big(\trans{\Leaves}f\big(\widehat{\h}(\x)\big), \y\big)\\
&\text{where}\hspace*{.5cm} \widehat{\h}(\x) =
\argmax{\h \in \Hset} \left\{\trans{\h} W \x \right\}~.
\end{aligned}
\end{equation}
Because we do not have access to the ground truth decisions {\em a
  priori}, this problem is a form of structured prediction with latent
variables. Drawing connection between latent variable structured
prediction and decision tree learning enables us to formulate an upper
bound on tree's empirical loss inspired by~\cite{TaskarGK03,
  TsochantaridisHJA04, YuJ09, NorouziFICML11}.

\comment{{
with a slight misuse of notation, loss function $\lossdec$ which associates each $\m$-bit binary
decision vector directly with a real valued cost:
\begin{equation}
\lossdec(\h, \y; \Leaves) = \loss(\trans{\Leaves}f(\h), \y)~.
\end{equation}
This loss $\lossdec$ encompasses the tree navigation function
$f(\cdot)$, and the leaf parameters $\Leaves$. Accordingly, empirical
loss can be expressed as a loss associated with a structured prediction
problem:
\begin{equation}
\begin{aligned}
\calL (\W, \Leaves ; \Dset) &=~ \frac{1}{\,\,n\,\,} \sum_{z=1}^n
\lossdec (\widehat{\h}_z, \y_z; \Leaves) \hspace*{1cm}\\
&\text{where}\hspace*{.5cm} \widehat{\h}_z =
\argmax{\h_z \in \Hset} \left\{\trans{\h}_z W \x_z \right\}~.
\end{aligned}
\end{equation}
Because we do not have access to the ground truth decisions {\em a priori}, this problem is a form of structured prediction
with latent variables. Drawing connection between latent variable structured prediction and decision tree learning enables us
to formulate an upper bound on tree's empirical loss inspired by~\cite{TaskarGK03, TsochantaridisHJA04, YuJ09, NorouziFICML11}.
}}

\comment{{
In the familiar structured prediction form, the

 to represent the potential binary
outcomes of the tests at the split nodes. Given an instance of $\h$, we can
determine based on binary bits in $\h$, which leaf will be reached.
Obviously, only the bits in $\h$ are relevant to our tree navigation task, which correspond to the
path from the root to the final leaf.

based on the outcome of each binary test, we decide how to navigate the
tree (whether to continue with the left or right child). Thus, new binary tests are selected based on

{
We now present an alternative representation of decision trees that is based
on latent decision variables associated with each node in the tree, and plays
a key part in enabling our method for jointly learning the split functions
of the trees.
}

Ignoring the hierarchical and sequential dependencies among
the binary tests in a decision tree,
We introduce an $\m$-bit latent decision variable
$\h \in \{-1, 1\}^\m$ to represent the potential binary
outcomes of the tests at the split nodes. Given an instance of $\h$, we can
determine based on binary bits in $\h$, which leaf will be reached.
Obviously, only the bits in $\h$ are relevant to our tree navigation task, which correspond to the
path from the root to the final leaf. The
function $f(\h): \{-1, 1\}^\m \to {\mathcal H}_{\m+1}$, where
${\mathcal H}_{\m+1}$ denotes a $1$-of-$(\m\!+\!1)$ encoding, maps
values of $\h$ to an indicator vector corresponding to a final
leaf. \figref{fig:fexample} depicts two examples of $f(\cdot)$ for a tree
of depth $\depth=2$ with $\m=3$ split nodes. Note that $\h$ can be an 
arbitrary $m$-bit vector,
and does not have to correspond to real binary tests on an input
$\x$. As becomes clear below, the introduction of the latent variable
$\h$, which can take values other than the outcomes of the binary
tests, plays an important role in our learning framework.
}}

}


\subsection{Upper Bound on Empirical Loss}

The upper bound on loss that we employ, given an input-output 
pair $(\x, \y)$, has the following form:
\comment{ To
overcome the difficulty of minimizing the discontinuous empirical
risk, we propose a surrogate piecewise smooth upper bound on the loss.
\begin{theorem}
For any pair $(\x, \y)$, the loss $\loss(\trans{\Leaves} f(\sign(\W
\x)), \y)$ is upper bounded by:
}
\begin{equation}
\begin{aligned}
& \mathbbm{1}(\trans{\w}\x < 0) \,\ell(\leaf_0, \y) + \mathbbm{1}(\trans{\w}\x \ge 0) \,\ell(\leaf_1, \y) \le \\
& \max\big(\!-\trans{\w}\x\! +\! \ell(\leaf_0, \y)~,~\trans{\w}\x + \ell(\leaf_1, \y) \big) -  \lvert \trans{\w}\x \rvert~,
\end{aligned}
\label{eq:upper-tree-loss}
\end{equation}
where $\lvert \trans{\w}\x \rvert$ denotes the absolute value of 
$\trans{\w}\x$.  To verify the bound, first suppose that $\trans{\w}\x < 0$. 
In this case, it is straightforward to show that the inequality reduces to
\begin{equation}
\ell(\leaf_0, \y) \, \le\, \max\big(\ell(\leaf_0, \y) ~,~ 
2 \,\trans{\w}\x +\ell(\leaf_1, \y) \big)~,
\label{eq:proof-tree-bound1}
\end{equation}
which holds trivially.  Conversely, when $\trans{\w}\x \ge 0$ the 
inequality reduces to 
\begin{equation}
\ell(\leaf_1, \y) \,\le \,
\max\big(\!-2 \,\trans{\w}\x + \ell(\leaf_0, \y) 
~,~ \ell(\leaf_1, \y) \big)~,
\label{eq:proof-tree-bound2}
\end{equation}
which is straightforward to validate.
Hence the inequality in~\eqref{eq:upper-tree-loss} holds.


Interestinly, while empirical loss in \eqref{eq:emp-risk-dtree} is 
invariant to $\lVert \w \rVert$, the bound in \eqref{eq:proof-tree-bound1} is not.
That is, for any real scalar $a > 0$, $\sign(a\trans{\w} \x)$ does not 
change with  $a$, and hence $\calL(\w,\leaf_{0,1}) = \calL(a\w,\leaf_{0,1})$.
Thus, while the loss on the LHS of \eqref{eq:upper-tree-loss} is 
scale-invariant, the upper bound on the RHS of \eqref{eq:upper-tree-loss} 
does depend on $\lVert \w \rVert$.  Indeed, like the soft-margin binary SVM 
formulation, and margin rescaling formulations of structural 
SVM~\cite{TsochantaridisHJA04}, the norm of $\w$ affects the interplay 
between the upper bound and empirical loss. In particular, as the scale 
of $\w$ increases, the upper bound becomes tighter and its optimization 
becomes more similar to a direct loss minimization.

More precisely, the upper bound becomes tighter as $\lVert\w\rVert$ 
increases. This is evident from the following inequality, which holds
for any real scalar $a > 1$:
\begin{equation}
\begin{aligned}
& \!\!\!\!\! 
\max\big(\!\!-\!\trans{\w}\x \! +\!  \ell(\leaf_0, \y), \, \trans{\w}\x 
\! +\!  \ell(\leaf_1, \y) \big) -  \lvert \trans{\w}\x \rvert ~ \ge\\
&~ \max\big(\!\!-\!a\trans{\w}\x\! +\! \ell(\leaf_0, \y),\, 
a\trans{\w}\x \!+\! \ell(\leaf_1, \y) \big) -  a\lvert \trans{\w}\x \rvert~.
\end{aligned}
\label{prop:co2-tightness}
\end{equation}
To verify the bound, as above, consider the sign of $\trans{\w}\x$.
When $\trans{\w}\x < 0$, inequality in \eqref{prop:co2-tightness} is
equivalent to
\begin{equation*}
\begin{aligned}
\max\big(\ell(\leaf_0, \y) ~,~ &2 \,\trans{\w}\x +\ell(\leaf_1, \y) \big) ~\ge\\
& 
\!\!\!\!
\!\!\!\!
\max \big(\ell(\leaf_0, \y) ~,~ 2 \,a\trans{\w}\x +\ell(\leaf_1, \y) \big)~.
\end{aligned}
\end{equation*}
Conversely, when $\trans{\w}\x \ge 0$, \eqref{prop:co2-tightness} is
equivalent to
\begin{equation*}
\begin{aligned}
\max\big(\!-2 \,\trans{\w}\x + &\ell(\leaf_0, \y) 
~,~ \ell(\leaf_1, \y) \big)~\ge\\
&
\!\!\!\!
\!\!\!\!
\max\big(\!-2\, a \trans{\w}\x + \ell(\leaf_0, \y) ~,~ \ell(\leaf_1, \y) \big)~.
\end{aligned}
\end{equation*}
Thus, as $\lVert \w \rVert$ increases the bound becomes tighter. In
the limit, as $\lVert \w \rVert$ becomes large, the loss terms
$\ell(\leaf_0, \y)$ and $\ell(\leaf_1, \y)$ become negligible compared
to the terms $-\trans{\w}\x$ and $\trans{\w}\x$, in which case the RHS
of \eqref{eq:upper-tree-loss} equals its LHS, except when
$\trans{\w}\x \approx 0$. Hence, for large $\lVert \w \rVert$, not
only the bound gets tight, but also it becomes less smooth and more
difficult to optimize in our nonconvex setting.

From the derivation above, and through experiments below, we observe
that when $\lVert \w \rVert$ is constrained, optimization converges to
better solutions that exhibit better generalization.  Summing over the
bounds for the training pairs, and restricting $\lVert \w \rVert$, we
obtain the surrogate objective we aim to optimize to find the decision
stump parameters:
\begin{eqnarray}
\begin{aligned}
&\mathrm{minimize}~~\calL'\left(\w, \leaf_{0,1} ; \Dset, \nu \right) \\
&~~~~\text{such that}~~~\lVert \w \rVert^2 \le \nu~,
\end{aligned}
\label{eq:upper-dtree-emploss}
\end{eqnarray}
where $\nu \in \Real^+$ is a regularization parameter, and $\calL'$ is 
the surrogate objective, \ie~the upper bound,
\begin{equation}
  \begin{aligned}
    \calL' & \left(\w, \leaf_{0,1} ; \Dset, \nu \right) ~\equiv \\
    & \sumxy \!\!\max\big(\!\!-\!\trans{\w}\x \!+\! \ell(\leaf_0, \y),\trans{\w}\x \!+\! \ell(\leaf_1, \y) \big) - \lvert \trans{\w}\x \rvert~.
  \end{aligned}
  \label{eq:upper-dtree-surrogate-objective}
\end{equation}
For all values of $\nu$, we have that $\calL' (\w, \leaf_{0,1} ;
\Dset, \nu) \ge \calL(\w, \leaf_{0,1} ; \Dset)$.  We find a suitable
$\nu$ via cross-validation. Instead of using the typical Lagrange form
for regularization, we employed hard constraints with similar behavior.

\subsection{Convex-Concave Optimization}

Minimizing the surrogate objective in \eqref{eq:upper-dtree-surrogate-objective}
entails nonconvex optimization.  While still challenging, it is important
that $\calL'\left(\w, \leaf_{0,1} ; \Dset, \nu \right)$ is better behaved 
than empirical loss.  It is piecewise smooth and convex-concave in $\w$, 
and the constraint on $\w$ defines a convex set. 
As a consequence, gradient-based optimization is applicable, although 
the surrogate objective is non-differentiable at isolated points. 
The objective also depends on the leaf parameters, $\leaf_0$ and $\leaf_1$, 
but only through the loss terms $\ell$, which we constrained to be convex 
in $\leaf$.  Therefore, for a fixed $\w$, 
it follows that $\calL'\left(\w, \leaf_{0,1} ;\, \Dset, \nu \right)$ 
is convex in $\leaf_0$ and $\leaf_1$.

The convex-concave nature of the surrogate objective allows us to use 
difference of convex (DC) programming, or the Convex-Concave Procedure
(CCCP)~\cite{YuilleR03}, a method for minimizing objective functions 
expressed as sum of a convex and a concave term. The CCCP has been 
employed by Felzenszwalb et al.~\yrcite{FelzenszwalbPAMI10} and Yu \&
Joachims~\yrcite{YuJ09} to optimize latent variable SVM models that
employ a similar convex-concave surrogate objective.

The Convex-Concave Procedure is an iterative method.  At each iteration
the concave term ($-\lvert \trans{\w}\x \rvert$ in our case) is replaced 
with its tangent plane at the current parameter estimate, to formulate a 
convex subproblem.  The parameters are updated with those that minimize 
the convex subproblem, and then the tangent plane is updated. 
Let $\iterold{\w}$ denote the estimate for $\w$ from the previous
CCCP iteration.  In the next iteration $\iterold{\w}$, $\leaf_0$,
and $\leaf_1$ are updated minimizing 
\begin{equation}
  \begin{aligned}
    \sumxy \Bigl(\max\big(\!\!-\!\trans{\w}\x & \!+\! \ell(\leaf_0, \y),\trans{\w}\x \!+\! \ell(\leaf_1, \y) \big) \\
    & - \sign(\trans{\iterold{\w}}\x)\, \trans{\w}\x \Bigr)~,
  \end{aligned}
\label{eq:upper-convex-dtree-emploss}
\end{equation}
such that $\lVert \w \rVert^2 \le \nu$. 

Note that $\iterold{\w}$ is constant during optimization of this 
CCCP subproblem.  In that case, the second term within the sum 
over training data in~\eqref{eq:upper-convex-dtree-emploss} just 
defines a hyperplane in the space of $\w$. 
The other (first) term within the sum entails maximization of a function
that is convex in $\w$, $\leaf_0$ and $\leaf_1$, since the maximum of 
two convex functions is convex. 
As a consequence, the objective of \eqref{eq:upper-convex-dtree-emploss} 
is convex.

We use stochastic subgradient descent to minimize
\eqref{eq:upper-convex-dtree-emploss}. After each subgradient update,
$\w$ is projected back into the feasible region.  For efficiency, we
do not wait for complete convergence of the convex subproblem within
CCCP.  Instead, $\iterold{\w}$ is updated after a fixed number of
epochs (denoted $\tau$) over the training dataset. The pseudocode for
the optimization procedure is outlined in Alg~\ref{alg:ccp-tree}.

\begin{algorithm}[t]
\caption{The convex-concave procedure for Continuous Optimization of
  Oblique (CO2) decision stumps that minimizes
  \eqref{eq:upper-dtree-emploss} to estimate ($\w$, $\leaf_0$,
  $\leaf_1$) given a training dataset $\Dset$, and a
    hyper-parameter $\nu$ that constrains the norm of $\w$}.
\begin{minipage}{\textwidth}
\begin{algorithmic}[1]
\vspace{.1cm}
\STATE Initialize $\w$ by a random univariate split
\STATE Estimate $\leaf_0$, and $\leaf_1$ based on $\w$ and $\Dset$
\WHILE{surrogate objective has not converged}
\STATE $\iterold{\w} \leftarrow {\w}$
\FOR{$t=1$ to $\tau$}
\STATE sample a pair $(\x, \y)$ at random from $\Dset$
  \STATE $s \leftarrow \sign(\iterold{\w}\x)$
  \IF{$-\trans{\w}\x + \ell(\leaf_0, \y) \ge \trans{\w}\x + \ell(\leaf_1, \y)$}
    \STATE ${\w} \leftarrow {\w} + \eta (1 + s) {\x}$
    \STATE ${\leaf_0} \leftarrow {\leaf_0} - \eta \,\partial \ell(\leaf_0, \y) / \partial \leaf$
  \ELSE
    \STATE ${\w} \leftarrow {\w} - \eta (1 - s) {\x}$
    \STATE ${\leaf_1} \leftarrow {\leaf_1} - \eta \,\partial \ell(\leaf_1, \y) / \partial \leaf$
  \ENDIF
  \IF{$\lVert \w \rVert_2^2 > \nu$}
    \STATE ${\w} \leftarrow \sqrt{\nu} \cdot {\w} / \lVert \w \rVert_2$
  \ENDIF
\ENDFOR
\ENDWHILE
\vspace{.1cm}
\end{algorithmic}
\end{minipage}
\label{alg:ccp-tree}
\end{algorithm}

In practice, we implement Alg~\ref{alg:ccp-tree} with several small
modifications.  Instead of estimating the gradients based on a single
data point, we use mini-batches of $100$ elements, and average their
gradients.  We also use a momentum term of $0.9$ to converge more
quickly. Finally, although a constant learning rate $\eta$ is used in
Alg~\ref{alg:ccp-tree}, we instead track the value of the surrogate
objective, and when it oscillates for more than a number of iterations
we reduce the learning rate.




\section{Implementation and Experimental Details}

In all tree construction methods considered here, we grow each
decision tree as deep as possible, until we reach a pure leaf.  
We exploit the bagging ensemble learning algorithm~\cite{bagging} 
to create the forest such that each tree is built by using a new 
data set sampled uniformly with replacement from the original dataset. 
In Random Forest, for finding each univariate split function we 
only consider a candidate set of size $q$ of random feature dimensions, 
where $q$ is the only hyper-parameter in our random forest 
implementation. We set the parameter $q$ by growing a forest of 
$1000$ trees and testing them on a hold-out validation set of 
size $20\%$ of the training set. Let $p$ denote the dimensionality
of the feature descriptors. We choose $q$ from the candidate set of
$\{p^{0.5}, p^{0.6}, p^{0.7}, p^{0.8}, p^{0.9}\}$ to accelerate
validation. Some previous work suggests the use of $q = \sqrt{p}$ as 
a heuristic~\cite{trevor2009elements}, which is included in the candidate 
set.

We use an OC1 implementation provided by the authors~\cite{oc1impl}.
We slightly modified the code to allow trees to grow to their fullest 
extent, removing hard-coded limits on tree depth and minimum 
examples for computing splits.  We also modified the initialization of 
OC1 optimization to match our initialization for CO2, whereby an optimal 
axis-aligned split on a subsampling of $q$ possible features is used.
Interestingly, we observed that both changes improve OC1's performance 
when building ensembles of multiple trees, {\em OC1 Forest}.  We use 
the default values provided by the authors for OC1 hyperparameters.

CO2 Forest has three hyper-parameters, namely, the regularization
parameter $\nu$, the initial learning rate $\eta$, and $q$, the size
of feature candidate set of which the best is selected to initialize
the CO2 optimization. Ideally, one may consider using different
regularizer parameters for different internal nodes of the tree, since
the number of available training data decreases as one descends the
tree.  However, we use the same regularizer and learning rate for all
of the nodes to keep hyper-parameter tuning simple.  We set $q$ as
selected by the random forest validation above.  We perform a grid
search over $\nu$ and $\eta$ to select the best hyper-parameters.

\section{Experiments}

\newlength\figh

Before presenting the classification results, we investigate the
impact of the hyper-parameter $\nu$ on our oblique decision trees.
\figref{fig:mnist-different-nu} depicts training and validation error
rates for the MNIST dataset for different values of $\nu \in \{0.1, 1,
10, 100\}$ and different tree depths. One can see that as the tree
depth increases, training error rate decreases monotonically.
However, validation error rate saturates at a certain depth, \eg~a
depth of $10$ for MNIST. Growing the trees deeper beyond this point,
either has no impact, or slightly hurts the performance.  From the plots it
appears that $\nu = 10$ exhibits the best training and validation
error rates. The difference between different values of $\nu$ seems to
be larger for validation error.

\begin{figure*}[t]
  \begin{center}
    \includegraphics[height=0.27\linewidth]{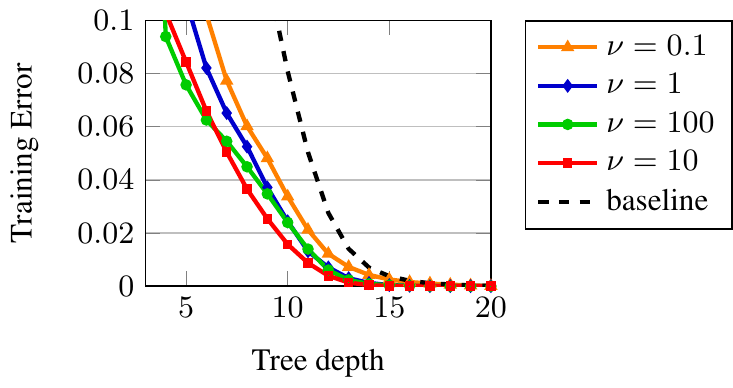}~
    \hspace{.2cm}~\includegraphics[height=0.27\linewidth]{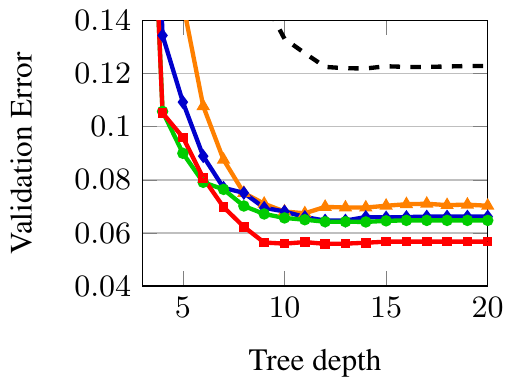}
  \end{center}
  \caption{The impact of hyper-parameter $\nu$ on MNIST training and
    validation error rates for CO2 decision trees. The dashed baseline
    represents univariate decision trees with no pruning.}
  \label{fig:mnist-different-nu}
\end{figure*}

As shown above in \eqref{prop:co2-tightness}, as $\nu$ increases
the upper bound becomes tighter. Thus, one might suspect that larger 
$\nu$ implies a better optimum and better training error rates.  
However, increasing $\nu$ not only tightens
the bound, but also makes the objective less smooth and harder to
optimize.  For MNIST, at $\nu = 10$ there appears to be a reasonable
balance between the tightness of the bound and the smoothness of the
objective. The hyper-parameter $\nu$ also acts as a
regularizer, contributing to the large gap in the validation
error rates. For completeness, we also include baseline results with
univariate decision trees. Clearly, the CO2 trees reach the same
training error rates as the baseline but at a smaller depth.  As seen
from the validation error rates, the CO2 trees achieve better
generalization too.

Classification results for tree ensembles are generally much better
than a single tree.  Here, we compare our {\em Continuously Optimized
  Oblique (CO2)} decision forest with {\em random
  forest}~\cite{Breiman01} and {\em OC1 forest}, forest built using
OC1 \cite{murthy1994}.  Results for random forest are obtained with
the implementation of the scikit-learn package~\cite{scikit-learn}.
Both of the baselines use information gain as the splitting criterion
for learning decision stumps.  We do not directly compare with other
types of classifiers, as our research concerns tree-based
techniques. Nevertheless, the reported results are often competitive
with the state-of-the-art.

\subsection{UCI multi-class benchmarks}

\begin{table*}
\renewcommand{\arraystretch}{1.3}
\begin{center}
\begin{tabular}{|c||cccc||ccc||ccc||ccc|}
\hline
  \multicolumn{5}{|c||}{}
& \multicolumn{9}{c|}{\bf Test error (\%) with different number of trees} \\
\cline{6-14}
\multicolumn{5}{|c||}{\bf Dataset Information}
& \multicolumn{3}{ c||}{\bf Random Forest}
& \multicolumn{3}{ c||}{\bf OC1 Forest}
& \multicolumn{3}{ c| }{\bf CO2 Forest}\\\cline{1-5}
\bf Name      & \bf \#Train     & \bf \#Test     & \bf \#Class & \bf Dim &
\bf 10 & \bf 30 & \bf 1000 & \bf 10 & \bf 30 & \bf 1000 & \bf 10 & \bf 30 &
\bf 1000\\
\hline
SatImage      & $4,435$     & $2,000$    & $6$  & $36$  &
10.1 & 9.4 & 8.9   &  10.1 & 9.9 & 9.5 &   9.6 & 9.1 & 8.9 \\
USPS          & $7,291$     & $2,007$    & $10$ & $256$ &
9.0 & 7.2 & 6.4    &  7.1 &  7.1 & 6.8  &   5.8 & 5.9 & 5.5 \\
Pendigits     & $7,494$     & $3,498$    & $10$ & $16$  &
3.9 & 3.3 & 3.5    &  3.2 &  2.2 & 2.3  &   1.8 & 1.7 & 1.7 \\
\hline
Letter        & $15,000$    & $5,000$    & $26$ & $16$  &
6.6 &  4.7 &  3.7 &  7.6 &  5.0 & 3.8  &   3.2 & 2.3 & 1.8 \\
Protein       & $17,766$    & $6,621$    & $3$  & $357$ &
39.9 & 35.5 & 30.9 & 39.1 & 34.6 & 30.8 &  33.8 & 31.2 & 30.3 \\
Connect4$^*$  & $55,000$  & $12,557$ & $3$ &$126$ &
18.9 & 17.4 & 16.2 & 
\multicolumn{3}{c||}{\multirow{1}{*}{N/A}} & 17.1 & 15.7 & 14.7 \\
\hline
MNIST         & $60,000$    & $10,000$   & $10$ & $784$ &
4.5 &  3.5 &  2.8 & \multicolumn{3}{c||}{\multirow{1}{*}{N/A}}
& 2.5 & 2.0 & 1.9\\
SensIT        & $78,823$    & $19,705$   & $3$  &$100$ &
15.5 & 14.0 & 13.4 &  
\multicolumn{3}{c||}{\multirow{1}{*}{N/A}}  & 14.1 & 13.0 & 12.5 \\
Covertype$^*$ & $500,000$ & $81,012$ & $7$ & $54$ &
3.2 &  2.8 &  2.6 &  
\multicolumn{3}{c||}{\multirow{1}{*}{N/A}} &    3.1 & 2.7 & 2.6 \\
\hline
\end{tabular}
\vspace*{.2cm}
\caption{Test error rates for forests with different number
  of trees on multi-class classification benchmarks. First few columns
  provide dataset information. \comment{including number of training
  and test points, number of classes, and feature
  dimensionality.} Test error rates (\%) for random forest,
  OC1 Forest, and CO2 Forest with $10$, $30$, and $1000$ trees are
  reported. For datasets marked with a star ``$^*$'' (\ie~Connect4 \&
  Covertype) we use our own training and test splits. As the number of
  training data points and feature dimensionality increase, OC1
  becomes prohibitively slow, so this method is not applicable to the
  datasets with high-dimensional data or large training sets.
\label{tab:multi-class-results}
}
\end{center}
\end{table*}

We conduct experiments on nine UCI multi-class benchmarks, namely, 
SatImage, USPS, Pendigits, Letter, Protein, Connect4, MNIST, SensIT, 
Covertype. \tabref{tab:multi-class-results} provides a summary
of the datasets, including the number of training and test points, the
number of class labels, and the feature dimensionality. We use the
training and test splits set by previous work, except for Connect4 and
Covertype. More details about the datasets, including references to
the corresponding publications can be found at the LIBSVM dataset
repository page~\cite{libsvmmulti}.

Test error rates for random forest, OC1 Forest, and CO2 Forest with
different numbers of trees ($10$, $30$, $1000$) are reported in
\tabref{tab:multi-class-results}. OC1 results are not presented on
some datasets, as the derivative-free coordinate descent method used
does not scale to large or high-dimensional datasets, \eg~requiring
more than $24$ hours to train a single tree on MNIST.  CO2 Forest
consistently outperforms random forest and OC1 Forest on all of the
datasets.  In some cases, \ie~Covertype, and SatImage, the improvement
is small, but in four of the datasets CO2 Forest with only $10$ trees
outperforms random forest with $1000$ trees.

For all methods, there is a large performance gain when the number 
of trees is increased from $10$ to $30$.  The marginal gain from 
$30$ to $1000$ trees is less significant, but still
notable. Finally, we also plot test error curves as a function of log
number of trees in \figref{fig:co2-forest-result}. CO2 Forest
outperforms random forest and OC1 by a large margin and in most cases
the marginal gain persists across different number of
trees. For some datasets, OC1 Forest outperforms random forest, 
but it consistently underperforms CO2 Forest.

\comment{
two image datasets that
are widely used as classification benchmarks, namely,
MNIST~\cite{MNIST} and CIFAR-10~\cite{alex}. The MNIST~\cite{MNIST}
digit dataset contains $60,000$ training and $10,000$ test images of
ten handwritten digits (0 to 9), each with $28 \!\times\!  28$ pixels.
Of the $60,000$ training images, we set aside $10,000$ for validation.
CIFAR-10~\cite{alex} comprises $50,000$ training and $10,000$ test
color images with $32 \!\times\! 32$ pixels.  Each image belongs to
one of 10 classes, namely airplane, automobile, bird, cat, deer, dog,
frog, horse, ship, and truck. Of the $50,000$ training images, we set
aside $10,000$ for validation. In both MNIST and CIFAR-10 we use raw
pixel values as features.

For the experimental results reported here we use Stable SGD (SSGD)
with fast loss augmented inference because according to our validation
experiments this method achieves the best accuracy at a reasonable
running time. Some interesting effects of parameter $\nu$ is depicted
in \figref{fig:effect-nu}. These effect are twofold. First, when $\nu$
increases for $d=10$ and $d=13$, training accuracy gets better up to a
point at which the function starts to become nonsmooth, and the
nonconvex optimizer does not change the parameters much from the
initialization. Hence, when $\nu$ gets large, we converge to the
initialization accuracy. Second, the $\nu$ parameter acts as a
regularizer and improves generalization power of the
tree. Interestingly, as bottom plots in \figref{fig:effect-nu} show,
as we increase $\nu$, more leaves in the tree become active. An active
leaf is a leaf to which a training exemplar is assigned. Because of
this effect of $\nu$ with different tree depth (\eg~$10$, $13$, and
$16$) we can achieve similar validation accuracy.

\definecolor{magenta}{rgb}{1,0,1}
\begin{figure*}
\footnotesize
\setlength\figh{3cm}

\begin{minipage}[ht]{0.33\linewidth}
\centering
\begin{tikzpicture}[scale=.9]
\begin{axis}[
xmode=log,
width=1.2\figh,
height=1\figh,
scale only axis,
xmin=1, xmax=1000,
ymin=0.85, ymax=1,
ylabel={Accuracy (\%)},
ymajorgrids,
legend style={at={(0.03,0.03)},anchor=south west,draw=black,fill=white}]

\addplot [
color=blue,
solid,
line width=1.25pt,
mark size=1.7pt,
mark=diamond*,
mark options={solid,fill=blue}
]
coordinates{(    1, 0.956205) ( 3.16, 0.966868) (   10, 0.973514) (31.62, 0.975619) (  100, 0.971645) (316.22, 0.962278) ( 1000, 0.944791)};
\addlegendentry{Training};

\addplot [
color=blue,
dashed,
line width=1.25pt,
mark size=0pt,
mark=*,
mark options={solid,fill=blue}
]
coordinates{(    1, 0.8602) ( 1000, 0.8602)};
\addlegendentry{Init. Train.};

\addplot [
color=red,
solid,
line width=1.25pt,
mark size=1.4pt,
mark=*,
mark options={solid,fill=red}
]
coordinates{(    1, 0.94484) ( 3.16, 0.951515) (   10, 0.95423) (31.62, 0.947305) (  100, 0.939655) (316.22, 0.91962) ( 1000, 0.896165)};
\addlegendentry{Validation};

\end{axis}
\end{tikzpicture}%

\vspace{.4cm}

\begin{tikzpicture}[scale=.9]
\begin{axis}[
xmode=log,
width=1.2\figh,
height=.7\figh,
scale only axis,
xmin=1, xmax=1000,
xlabel={Regularization parameter $\nu$ ($\log$)},
ymin=0, ymax=4000,
ylabel={Num. active leaves},
ymajorgrids,
legend style={at={(0.03,0.03)},anchor=south west,draw=black,fill=white}]
\addplot [
color=green!60!black,
solid,
line width=1.25pt,
mark size=1.4pt,
mark=*,
mark options={solid,fill=green!60!black}
]
coordinates{(    1,   296) ( 3.16,   413) (   10,   477) (31.62,   580) (  100,   635) (316.22,   730) ( 1000,   827)};

\addplot [
color=green!60!black,
dashed,
line width=1.25pt,
mark size=0pt,
mark=*,
mark options={solid,fill=blue}
]
coordinates{(    1, 924) ( 1000, 924)};

\end{axis}
\end{tikzpicture}%

Tree depth $d$ =10
\end{minipage}~\hspace*{.3cm}~\begin{minipage}[ht]{0.30\linewidth}  
\centering
\hspace*{.15cm}\begin{tikzpicture}[scale=.9]
\begin{axis}[
xmode=log,
width=1.2\figh,
height=\figh,
scale only axis,
xmin=1, xmax=1000,
ymin=0.85, ymax=1,
ymajorgrids,
legend style={at={(0.03,0.03)},anchor=south west,draw=black,fill=white}]

\addplot [
color=blue,
solid,
line width=1.25pt,
mark size=1.7pt,
mark=diamond*,
mark options={solid,fill=blue}
]
coordinates{(    1, 0.959409) ( 3.16, 0.972267) (   10, 0.979218) (31.62, 0.981629) (  100, 0.98043) (316.22, 0.975719) ( 1000, 0.974383)};
\addlegendentry{Training};

\addplot [
color=blue,
dashed,
line width=1.25pt,
mark size=0pt,
mark=*,
mark options={solid,fill=blue}
]
coordinates{(    1, 0.971) ( 1000, 0.971)};
\addlegendentry{Init. Train.};

\addplot [
color=red,
solid,
line width=1.25pt,
mark size=1.4pt,
mark=*,
mark options={solid,fill=red}
]
coordinates{(    1, 0.94761) ( 3.16, 0.95448) (   10, 0.95234) (31.62, 0.94842) (  100, 0.92436) (316.22, 0.877055) ( 1000, 0.85214)};
\addlegendentry{Validation};

\end{axis}
\end{tikzpicture}%

\vspace{.4cm}

\begin{tikzpicture}[scale=.9]
\begin{axis}[
xmode=log,
width=1.2\figh,
height=.7\figh,
scale only axis,
xmin=1, xmax=1000,
xlabel={Regularization parameter $\nu$ ($\log$)},
ymin=0, ymax=4000,
ymajorgrids,
legend style={at={(0.03,0.03)},anchor=south west,draw=black,fill=white}]
\addplot [
color=green!60!black,
solid,
line width=1.25pt,
mark size=1.4pt,
mark=*,
mark options={solid,fill=green!60!black}
]
coordinates{(    1,   393) ( 3.16,   612) (   10,   900) (31.62,  1166) (  100,  1493) (316.22,  2262) ( 1000,  2485)};

\addplot [
color=green!60!black,
dashed,
line width=1.25pt,
mark size=0pt,
mark=*,
mark options={solid,fill=blue}
]
coordinates{(    1, 2947) ( 1000, 2947)};

\end{axis}
\end{tikzpicture}%

Tree depth $d$ =13
\end{minipage}~\hspace*{.3cm}~\begin{minipage}[ht]{0.30\linewidth}  
\centering
\hspace*{.15cm}\begin{tikzpicture}[scale=.9]
\begin{axis}[
xmode=log,
width=1.2\figh,
height=\figh,
scale only axis,
xmin=1, xmax=1000,
ymin=0.85, ymax=1,
ymajorgrids,
legend style={at={(0.03,0.03)},anchor=south west,draw=black,fill=white}]

\addplot [
color=blue,
solid,
line width=1.25pt,
mark size=1.7pt,
mark=diamond*,
mark options={solid,fill=blue}
]
coordinates{(    1, 0.958978) ( 3.16, 0.972395) (   10, 0.98039) (31.62, 0.983175) (  100, 0.98391) (316.22, 0.987619) ( 1000, 0.992256)};
\addlegendentry{Training};

\addplot [
color=blue,
dashed,
line width=1.25pt,
mark size=0pt,
mark=*,
mark options={solid,fill=blue}
]
coordinates{(    1, 0.99858) ( 1000, 0.99858)};
\addlegendentry{Init. Train.};

\addplot [
color=red,
solid,
line width=1.25pt,
mark size=1.4pt,
mark=*,
mark options={solid,fill=red}
]
coordinates{(    1, 0.94741) ( 3.16, 0.95636) (   10, 0.953505) (31.62, 0.941595) (  100, 0.912925) (316.22, 0.855435) ( 1000, 0.82951)};
\addlegendentry{Validation};

\end{axis}
\end{tikzpicture}%

\vspace{.4cm}
\begin{tikzpicture}[scale=.9]
\begin{axis}[
xmode=log,
width=1.2\figh,
height=.7\figh,
scale only axis,
xmin=1, xmax=1000,
xlabel={Regularization parameter $\nu$ ($\log$)},
ymin=0, ymax=4000,
ymajorgrids,
legend style={at={(0.03,0.03)},anchor=south west,draw=black,fill=white}]
\addplot [
color=green!60!black,
solid,
line width=1.25pt,
mark size=1.4pt,
mark=*,
mark options={solid,fill=green!60!black}
]
coordinates{(    1,   404) ( 3.16,   561) (   10,   986) (31.62,  1562) (  100,  2175) (316.22,  3051) ( 1000,  3402)};

\addplot [
color=green!60!black,
dashed,
line width=1.25pt,
mark size=0pt,
mark=*,
mark options={solid,fill=blue}
]
coordinates{(    1, 3726) ( 1000, 3726)};

\end{axis}
\end{tikzpicture}%

Tree depth $d$ =16
\end{minipage}%
\caption{Validation results on MNIST validation set.}
\label{fig:effect-nu}
\end{figure*}
}

For pre-processing, the datasets are scaled so that either the
feature dimensions are in the range of $[0, 1]$, or they have a zero
mean and a unit standard deviation.  For CO2 Forest, we select $\nu$
from the set $\{0.1, 1, 4, 10, 43, 100\}$, and $\eta$ from the set
$\{.03, .01, .003\}$. A validation of $30$ decision trees is performed
over $18$ entries of the grid of $(\nu, \eta)$.

\newcommand*{\forestplotwidth}{5.1cm}
\newcommand*{\forestplotspace}{}

\begin{figure*}[t]
  \begin{center}
    \begin{tabular}{@{}c@{~~~~~}c@{~~~~~}c@{}}
      \forestplotspace~\includegraphics[width=\forestplotwidth]{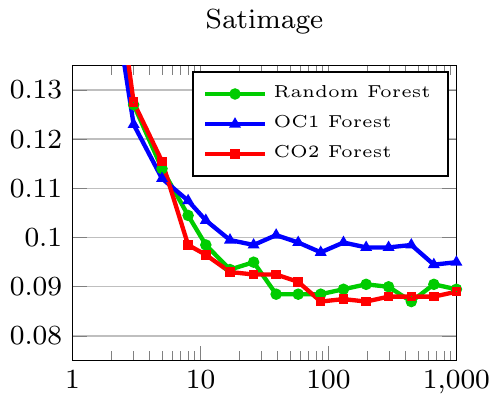} &
      \forestplotspace~\includegraphics[width=\forestplotwidth]{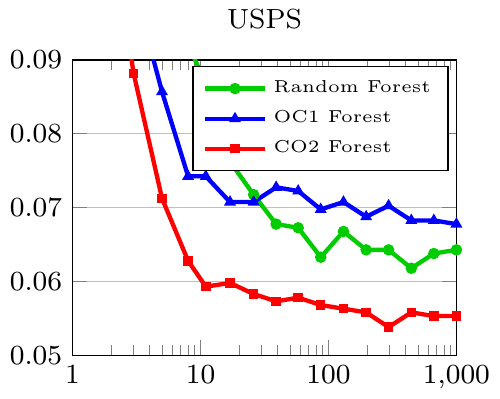} &
      \forestplotspace~\includegraphics[width=\forestplotwidth]{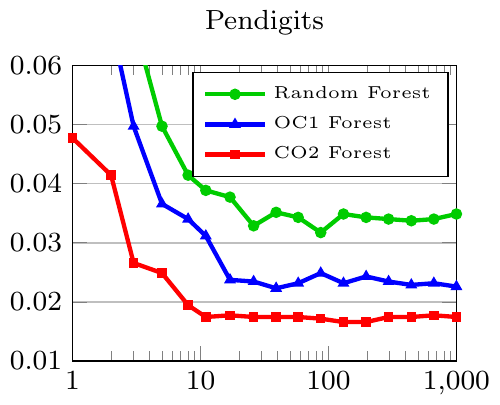} \\
      \forestplotspace~\includegraphics[width=\forestplotwidth]{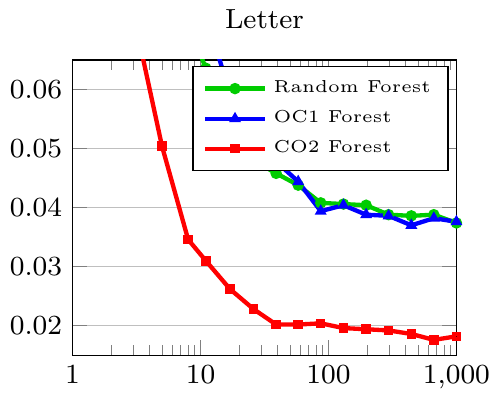} &
      \forestplotspace~\includegraphics[width=\forestplotwidth]{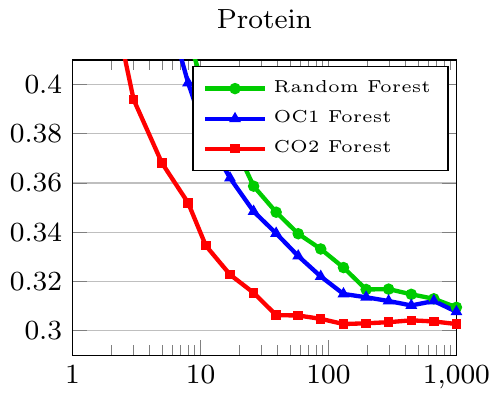} &
      \forestplotspace~\includegraphics[width=\forestplotwidth]{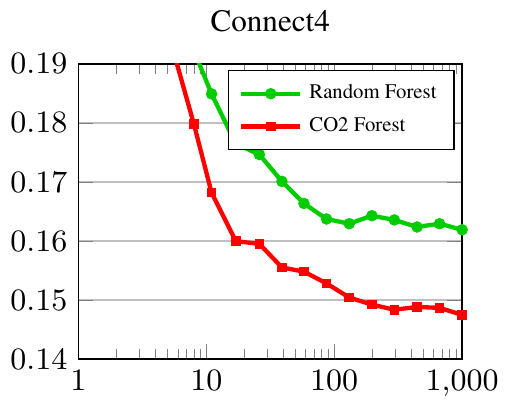} \\
      \forestplotspace~\includegraphics[width=\forestplotwidth]{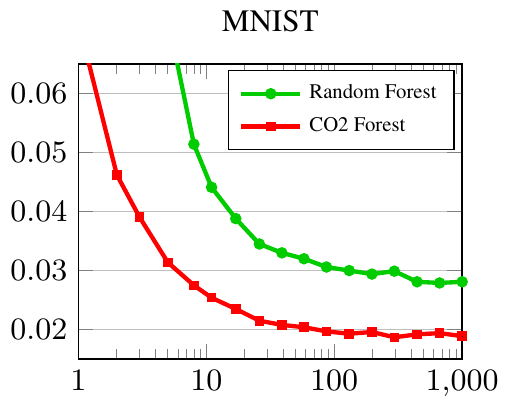} &
      \forestplotspace~\includegraphics[width=\forestplotwidth]{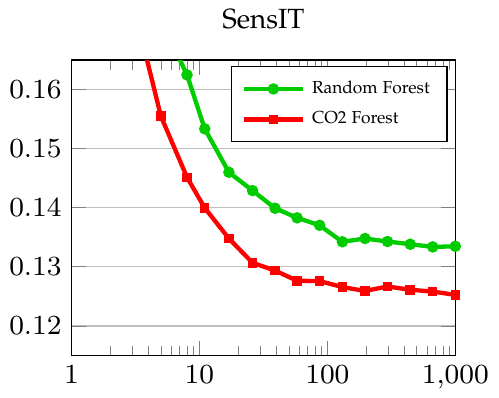} &
      \forestplotspace~\includegraphics[width=\forestplotwidth]{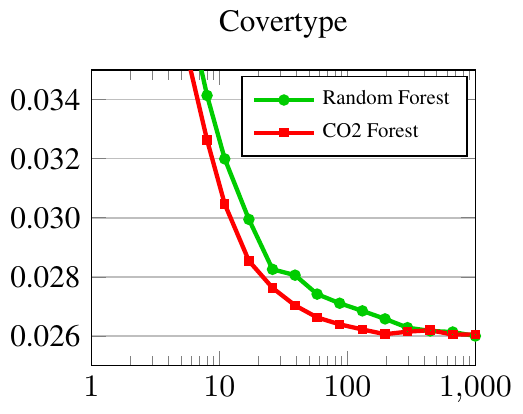} \\
    \end{tabular}
  \end{center}
  \caption{Test error curves for Random Forest and OC1 Forest \vs~CO2
    Forest as a function of (log) number of trees on the multi-class
    classification benchmarks. On the last four datasets, OC1
      implementation is prohibitively slow, hence not applicable.}
  \label{fig:co2-forest-result}
\end{figure*}

\subsection{Labeled Faces in the Wild (LFW)}

\newcommand{\selectedimagewidth}{1.67cm}
\begin{figure*}
\begin{center}
\begin{tabular}{@{}c@{~}c@{~}c@{~}c@{~}c@{\hspace*{.5cm}}c@{~}c@{~}c@{~}c@{~}c@{}}
  \scriptsize Input & 
  \scriptsize Ground truth &
  \scriptsize Axis-aligned &
  \scriptsize Two-probe &
  \scriptsize CO2 &
  \scriptsize Input & 
  \scriptsize Ground truth &
  \scriptsize Axis-aligned &
  \scriptsize Two-probe &
  \scriptsize CO2 \\[.1cm]
    \includegraphics[width=\selectedimagewidth]{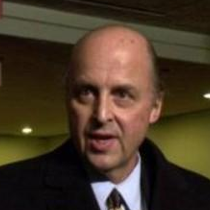} &
    \includegraphics[width=\selectedimagewidth]{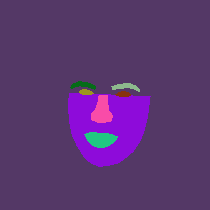} &
    \includegraphics[width=\selectedimagewidth]{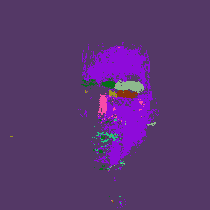} &
    \includegraphics[width=\selectedimagewidth]{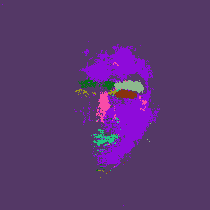} &
    \includegraphics[width=\selectedimagewidth]{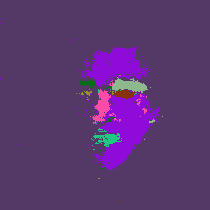} &

    \includegraphics[width=\selectedimagewidth]{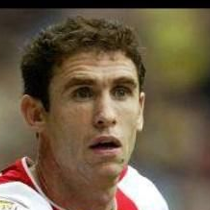} &
    \includegraphics[width=\selectedimagewidth]{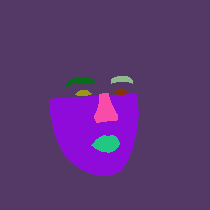} &
    \includegraphics[width=\selectedimagewidth]{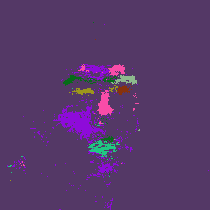} &
    \includegraphics[width=\selectedimagewidth]{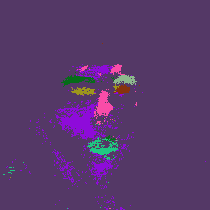} &
    \includegraphics[width=\selectedimagewidth]{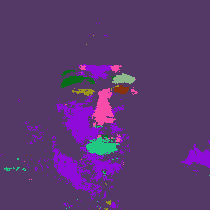} \\

    \includegraphics[width=\selectedimagewidth]{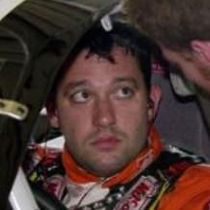} &
    \includegraphics[width=\selectedimagewidth]{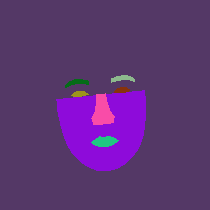} &
    \includegraphics[width=\selectedimagewidth]{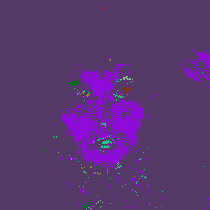} &
    \includegraphics[width=\selectedimagewidth]{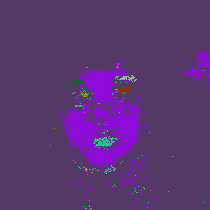} &
    \includegraphics[width=\selectedimagewidth]{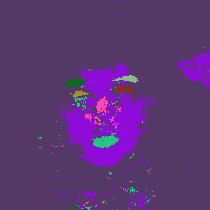} &

    \includegraphics[width=\selectedimagewidth]{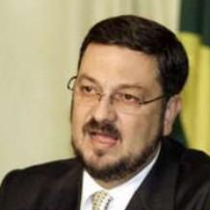} &
    \includegraphics[width=\selectedimagewidth]{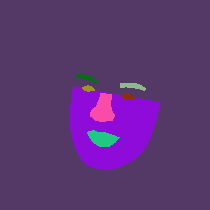} &
    \includegraphics[width=\selectedimagewidth]{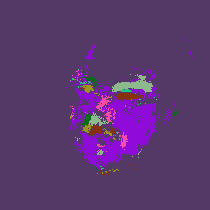} &
    \includegraphics[width=\selectedimagewidth]{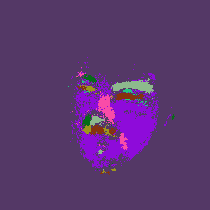} &
    \includegraphics[width=\selectedimagewidth]{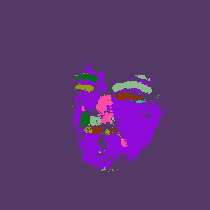} \\

    \includegraphics[width=\selectedimagewidth]{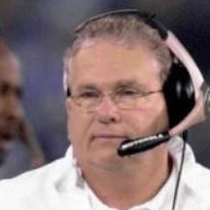} &
    \includegraphics[width=\selectedimagewidth]{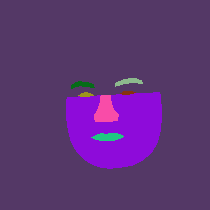} &
    \includegraphics[width=\selectedimagewidth]{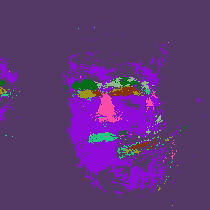} &
    \includegraphics[width=\selectedimagewidth]{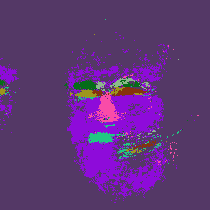} &
    \includegraphics[width=\selectedimagewidth]{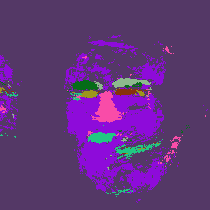} &

    \includegraphics[width=\selectedimagewidth]{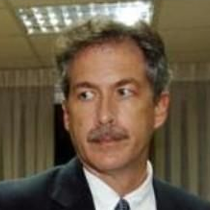} &
    \includegraphics[width=\selectedimagewidth]{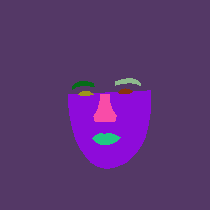} &
    \includegraphics[width=\selectedimagewidth]{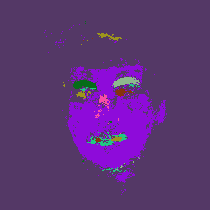} &
    \includegraphics[width=\selectedimagewidth]{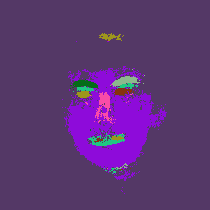} &
    \includegraphics[width=\selectedimagewidth]{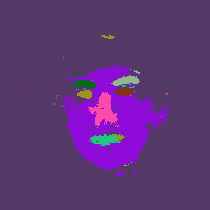} \\

    \includegraphics[width=\selectedimagewidth]{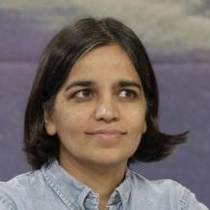} &
    \includegraphics[width=\selectedimagewidth]{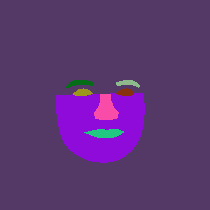} &
    \includegraphics[width=\selectedimagewidth]{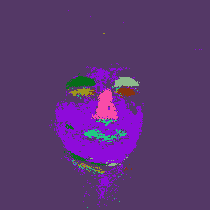} &
    \includegraphics[width=\selectedimagewidth]{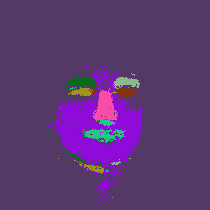} &
    \includegraphics[width=\selectedimagewidth]{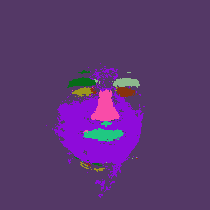} &

    \includegraphics[width=\selectedimagewidth]{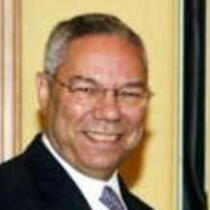} &
    \includegraphics[width=\selectedimagewidth]{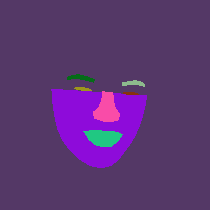} &
    \includegraphics[width=\selectedimagewidth]{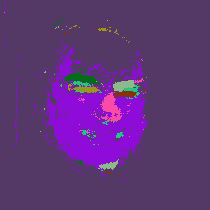} &
    \includegraphics[width=\selectedimagewidth]{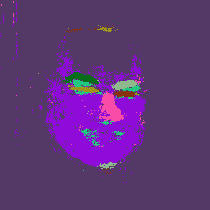} &
    \includegraphics[width=\selectedimagewidth]{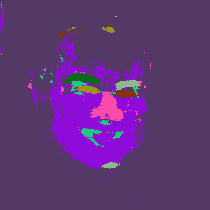} \\

    \includegraphics[width=\selectedimagewidth]{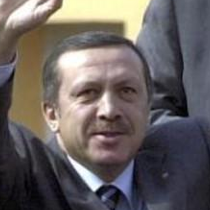} &
    \includegraphics[width=\selectedimagewidth]{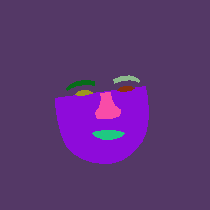} &
    \includegraphics[width=\selectedimagewidth]{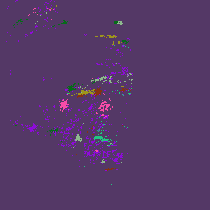} &
    \includegraphics[width=\selectedimagewidth]{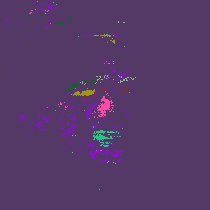} &
    \includegraphics[width=\selectedimagewidth]{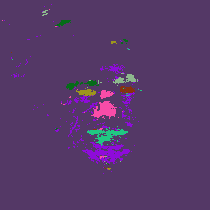} &

    \includegraphics[width=\selectedimagewidth]{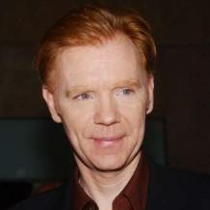} &
    \includegraphics[width=\selectedimagewidth]{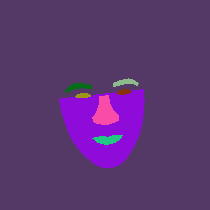} &
    \includegraphics[width=\selectedimagewidth]{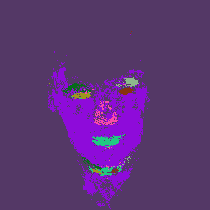} &
    \includegraphics[width=\selectedimagewidth]{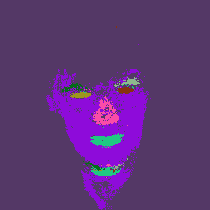} &
    \includegraphics[width=\selectedimagewidth]{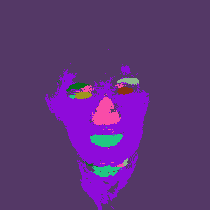} \\
\end{tabular}
\vspace*{-.1cm}
\end{center}
\caption{ \label{fig:selected} The above shows side-by-side
  comparisons of classification results on images from the test set.
  From left to right, these are the input image, the ground truth
  labels, the outputs of Random Forest, Two-probe Forest, and
    CO2 Forest.  The first column depicts images chosen to
  show a spread of different segmentation qualities with the first
  three the median image by Jaccard score for each method, and the
  last two, examples with highest mean and variance in scores across
  methods.  The second column shows five random examples.  }
\vspace*{-.2cm}
\end{figure*}

As a large-scale experiment, we consider the task of segmenting face
parts based on the Labeled Faces in the Wild (LFW)
dataset \cite{huang2007-lfw}.
We seek to label image pixels that belong to one of the following $7$
face parts: lower face, nose, mouth, and the left and right eyes and
eyebrows.  These parts should be differentiated from the background,
which provides a total of $8$ class labels.  To address this task,
decision trees are trained on $31 \times 31$ image sub-windows to
predict the label of the center pixel.  Each $31 \times 31$ window
with three RGB channels is vectorized to create an input in
$\Real^{2883}$.  We ignore part labels for a $15$-pixel border around
each image at both training and test time.

To train each tree, we subsample $256,\!000$ sub-windows from training
images.  We then normalize the pixels of each window to be of unit
norm and variance across the training set.  The same transformation is
applied to input windows at 
test time using the normalization parameters calculated on the
training set.  To correct for the class label
imbalance, like~\cite{kontschieder2013-geodesic-forests}, we 
subsample training windows so that each label has an equal number 
of training examples.  At test time, we reweight the class 
label probabilities given by the inverse of the factor that each label 
was undersampled or oversampled during training.

Other than the random forest baseline, we also train decision trees
and forests using split functions that compare two features (or ``probes''), 
where the choice of features comes from finding the optimal pair 
of features out of a large number of sampled pairs. This method 
produces decision forests analogous to \cite{ShottonPami13}.  We call this 
baseline {\em Two-probe Forest}. The same technique can be used to 
generate split functions with several features, but we found that using 
only two features produces the best accuracy on the validation set.

Because of the class label imbalance in LFW, classification accuracy
is a poor measure of the segmentation quality.  A more informative
performance measure, also used in the PASCAL VOC challenge, is the
class-average Jaccard score.  We report Jaccard scores for the
baselines {\em vs.} CO2 Forest in~\tabref{tab:jaccards}. It is clear
that Two-probe Forest outperforms random forest, and CO2 Forest
outperforms both of the baselines considerably. The superiority of CO2
Forest is consistent in~\figref{fig:lfw_limits}, where Jaccard scores
are depicted for forests with fixed tree depths, and forests with
different number of trees.  The Jaccard score is calculated for each
class label, against all of the other classes, as $100 \cdot tp / (tp
+ fp + fn)$.  The average of this quantity over classes is reported
here. The test set comprises $250$ randomly chosen images.

\begin{table}
\begin{center}
\renewcommand{\arraystretch}{1.3}
\begin{tabular}{|ccc|}
  \hline
  \textbf{Technique} & \textbf{16 trees} & \textbf{32 trees} \\
  \hline
  Random Forest        & 32.28 & 34.61 \\
  Two-probe Forest   & 36.03 & 38.61 \\
  CO2 Forest                 & 40.33 & 42.55 \\
  \hline
\end{tabular}
\end{center} 
\caption{ \label{tab:jaccards}
  Test Jaccard scores comparing CO2 Forest to baseline forests on the
  Labeled Faces in the Wild (LFW) dataset.}
\end{table}

We use the Jaccard score to select the CO2 hyperparameters $\nu$ 
and $\eta$. We perform grid search over 
$\eta \in \{10^{-5}, 10^{-4}, 3 \cdot 10^{-4}, 6 \cdot 10^{-4}, 0.001,
0.003\}$ and $\nu \in \{0.1, 1, 4, 10, 43, 100\}$.  We compare the
scores for 16 trees on a held-out validation set of $100$ images.  
The choice of $\eta = 10^{-4}$ and $\nu = 1$ achieves the highest
validation Jaccard score of $41.87$, and are used in the final
experiments.

We note that some other tree-like structures 
\cite{shotton2013-decision-jungles} and more sophisticated Computer 
Vision systems built for face segmentation \cite{smith2013-exemplar-faces} 
achieve better segmentation accuracy on LFW. However, our models use only 
raw pixel values, and our goal was to compare CO2 Forest against 
forest baselines.

\begin{figure}
  \begin{center}
    \includegraphics[height = 0.38\linewidth]{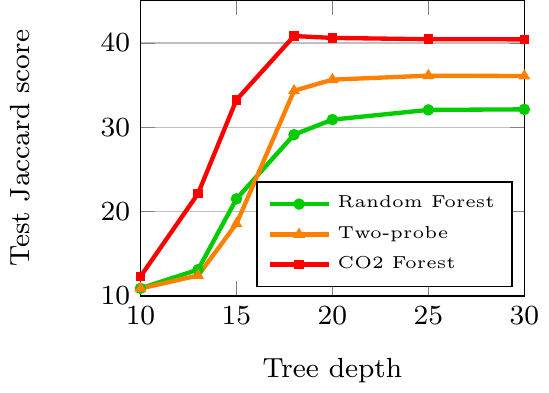}
    \nolinebreak[4]
    \includegraphics[height = 0.38\linewidth]{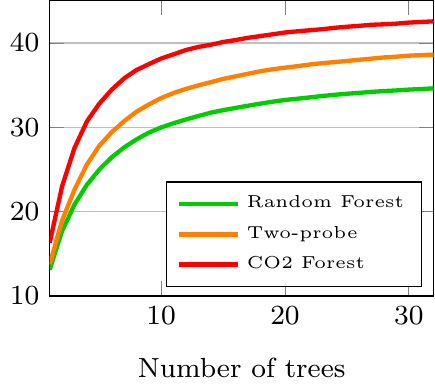}
  \end{center}

  \caption{Test Jaccard scores on LFW for (left) forests of
  $16$ trees with different tree depth constraint from $10$ to $30$ (right) 
forests with different number of trees from $1$ to $32$.}
\label{fig:lfw_limits}
\end{figure}

\section{Conclusion}

We present Continuously Optimized Oblique (CO2) Forest, a new variant
of random forest that uses oblique split functions.  Even though the
information gain criterion used for inducing decision trees is
discontinuous and hard to optimize, we propose a continuous upper
bound on the information gain objective.  We leverage this bound to
optimize oblique decision tree ensembles, which achieve a large
improvement on classification benchmarks over a random forest baseline
and previous methods of constructing oblique decision trees.  In
contrast to OC1 trees, our method scales to problems with
high-dimensional inputs and large training sets, which are
commonplace in Computer Vision and Machine Learning.
Our framework is straightforward to generalize to other
tasks, such as regression or structured prediction, as the upper bound
is general and applies to any form of convex loss function.


\ifCLASSOPTIONcaptionsoff
  \newpage
\fi



%


%

\bibliographystyle{ieee}
\bibliography{co2}




\end{document}